\documentclass[runningheads]{llncs}
\usepackage[T1]{fontenc}
\usepackage{graphicx}
\usepackage{amsmath} 
\usepackage{amsfonts} 
\usepackage{xcolor}
\usepackage{hyperref} 

\begin{document}
\title{Disentangled Diffusion Autoencoder for Harmonization of Multi-site Neuroimaging Data}
\titlerunning{DDAE for Harmonization of Multi-site Neuroimaging Data}
\authorrunning{Ijishakin et al.}
\author{Ayodeji Ijishakin\inst{1}\thanks{Equal Contributions} \and Ana Lawry Aguila\inst{1}$^{\star}$ \and Elizabeth Levitis\inst{1} \and Ahmed Abdulaal\inst{1} \and Andre Altmann\inst{1} \and James Cole\inst{1}}
\institute{Centre for Medical Image Computing, Department of Computer Science, University College London, UK
 \\\email{Corresponding Email: ayodeji.ijishakin.21@ucl.ac.uk}} 
\maketitle         
\begin{abstract}
Combining neuroimaging datasets from multiple sites and scanners can help increase statistical power and thus provide greater insight into subtle neuroanatomical effects. However, site-specific effects pose a challenge by potentially obscuring the biological signal and introducing unwanted variance. Existing harmonization techniques, which use statistical models to remove such effects, have been shown to incompletely remove site effects while also failing to preserve biological variability. More recently, generative models using GANs or autoencoder-based approaches, have been proposed for site adjustment. However, such methods are known for instability during training or blurry image generation. In recent years, diffusion models have become increasingly popular for their ability to generate high-quality synthetic images. In this work, we introduce the disentangled diffusion autoencoder (DDAE), a novel diffusion model designed for controlling specific aspects of an image. We apply the DDAE to the task of harmonizing MR images by generating high-quality site-adjusted images that preserve biological variability. We use data from 7 different sites and demonstrate the DDAE's superiority in generating high-resolution, harmonized 2D MR images over previous approaches. As far as we are aware, this work marks the first diffusion-based model for site adjustment of neuroimaging data.
\end{abstract}

\section{Introduction and related work}
Pooling data from multiple sites in large-scale neuroimaging studies increases statistical power, thus enabling the detection of more subtle disease effects. However, aggregating data across different sites, scanners, or acquisition parameters introduces unwanted variance which can obscure the signal of interest. As such, several data harmonization methods to remove site effects have been proposed. ComBat \cite{Johnson2006} is a popular data harmonization tool that was initially introduced for genomics research and has since been adapted for harmonizing neuroimaging data. It uses an empirical Bayes approach to adjust for linear site effects and has been widely applied to multiple neuroimaging modalities \cite{Fortin2018,Fortin2017}. To preserve known sources of biological variance, biological covariates are integrated into the ComBat model.
\begin{figure*}[t]
\centering
    \includegraphics[scale=0.2]{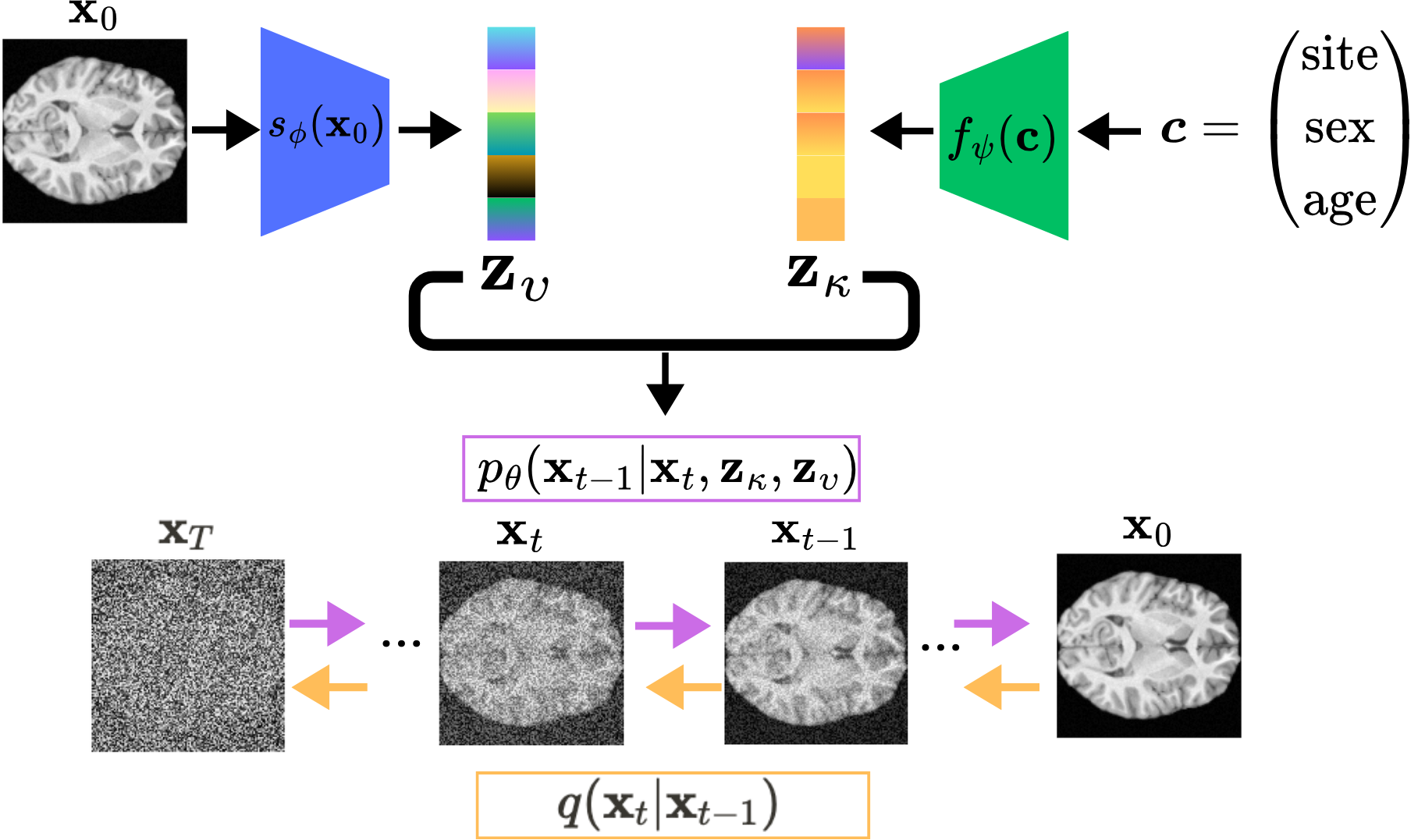}
\caption{The disentangled diffusion autoencoder. Our image $\mathbf{x}_{0}$ goes through the unknown-variance encoder $s_{\phi}(\mathbf{x}_{0})$ to produce $\mathbf{z}_{\upsilon}$. Next, the known-variance encoder $f_{\psi}(\boldsymbol{c})$ produces $\mathbf{z}_{\kappa}$ conditional on age, sex and site. Our forward process $q(\mathbf{x}_{t}| \mathbf{x}_{0})$ noises our data $\mathbf{x}_{0}$ to produce $\mathbf{x}_{t}$. After which our reverse process $p_{\theta}(\mathbf{x}_{t}|\mathbf{z}_{\kappa}$, $\mathbf{z}_{\upsilon})$ then recovers our data conditional on $\mathbf{z}_{\kappa}$ and $\mathbf{z}_{\upsilon}$.} 
\label{ModelArch}
\end{figure*} 
However, there are several limitations to ComBat-based approaches. For example, it has been demonstrated that ComBat falls short of eliminating site effects while also failing to preserve biological variability and variance in the data originating from unknown sources \cite{Mostafa20}. To overcome these limitations, Kia et al. use Hierarchical Bayesian Regression (HBR) to adjust for batch effects from multi-site data within a normative modelling framework \cite{Mostafa20}. However, the HBR framework is designed for normative modelling, and the batch removal element cannot be separated for use in other applications. 

Recently, deep generative models have been applied to the task of generating neuroimaging data with specific properties. For example, Lawry Aguila et al. (2022) \cite{LawryAguila2022}  use a conditional Variational Autoencoder (cVAE) to control for confounding variables, in an approach that could readily be adapted for the removal of multi-site effects. However, VAEs are known to generate blurry reconstructed images. Alternatively, focusing specifically on multi-site harmonization, Zhao et al. (2023) \cite{Zhao2023} propose a cycle-consistent Adversarial Autoencoder to generate data adjusted for site effects, showing superior performance over ComBat variants in both site effects removal and preservation of biological variability. However, this model is designed for imaging data mapped onto a spherical surface and is not currently suitable for raw brain images. The most relevant deep learning-based harmonization method is the Style-Encoding GAN, which uses a Generative Adversarial Network (GAN) with a style encoder to generate site-invariant MR images \cite{Liu2021}. 

Diffusion models have been shown to outperform GANs in image synthesis, consistently achieving improved image quality across a range of datasets \cite{Dhariwal2021,Rombach2022}. They can capture more expressive representations of complex data than previous generative models, by modelling the data distribution via a set of fixed Gaussian encoding steps and learnable reverse steps. Diffusion models have recently been applied to MR data for disease classification and data generation \cite{Ijishakin2023,Pinaya2022}.

In this work, we present a novel diffusion model, called the disentangled diffusion autoencoder, which incorporates conditional variables into the modelling process to control specific aspects of the image generation process. We showcase the practical utility of our model by applying it to the task of generating harmonized MR images. We compare our proposed model to ComBat, a cVAE, and Style-Encoding GAN \cite{Liu2021} on the task of harmonizing images from 7 different neuroimaging sites. Our experiments demonstrate that our model generates high quality images, removes site effects, and preserves biological variability from known and unknown sources, whilst other approaches do not consistently perform well across all of these areas. 

\section{Background}
Diffusion models are hierarchical latent variable models, where our latent variables, $\mathbf{x}_{1}, ...,\mathbf{x}_{T}$ are time dependent and share the same dimension as our data $\mathbf{x}_{0} \sim q(\mathbf{x}_{0})$. To learn our data distribution, $q(\mathbf{x}_{0})$, we express the joint distribution over all latents as a Markov chain, starting at a prior distribution  $p(\mathbf{x}_{T}) = \mathcal{N}(\mathbf{x}_{T}; 0, \textbf{I})$ and ending at our model distribution $p_{\theta}(\mathbf{x}_{0})$. 
\begin{equation}
     p_{\theta}(\mathbf{x}_{0:T}) := p(\mathbf{x}_{T})\prod_{t=1}^{T}p_{\theta}^{(t)}(\mathbf{x}_{t-1}| \mathbf{x}_{t})
\end{equation}
This joint distribution, $p_{\theta}(\mathbf{x}_{0:T})$, is known as the \textit{reverse process}, and it allows us to recover the probability of our data $p_{\theta}(\mathbf{x}_{0})$ by approximating $p_{\theta}(\mathbf{x}_{0}) = \int p_{\theta}(\mathbf{x}_{0:T})d\mathbf{x}_{1:T}$. Our approximate posterior, $q(\mathbf{x}_{1:T}|\mathbf{x}_{0})$, is referred to as the \textit{forward process}. The \textit{forward process} iteratively corrupts our data through the addition of Gaussian noise, which simulates a diffusion process. It is pre-defined as another Markov chain of the form: 
\begin{equation}
    q(\mathbf{x}_{1:T}|\mathbf{x}_{0}) := \prod^{T}_{t=1} q(\mathbf{x}_{t}|\mathbf{x}_{t-1}), \quad  q(\mathbf{x}_{t}|\mathbf{x}_{t-1}) := \mathcal{N}(\mathbf{x}_{t};\sqrt{1-\beta_{t}} \mathbf{x}_{t-1}, \beta_{t} \textbf{I})
\end{equation}
Where $\beta_{1}, ..., \beta_{T}$ are scalar values that control the variance at each time step, $t$. As such, diffusion models are a two-step procedure, our data $\mathbf{x}_{0} \sim q(\mathbf{x}_{0})$ is corrupted by the forward diffusion process, $q(\mathbf{x}_{1:T}|\mathbf{x}_{0})$, and then denoised by the reverse process $p_{\theta}(\mathbf{x}_{0:T})$ to obtain our original density.  

A variational lower bound on log-likelihood may be maximised to learn our model distribution $p_{\theta}({\mathbf{x}}_{0})$:
\begin{equation}
    \label{lower-bound}
    \mathbb{E} \left[ \log p_{\theta}(\mathbf{x}_{0}) \right] \geq \mathbb{E}_{q} \left[ \log \frac{p_{\theta}(\mathbf{x}_{0:T})}{q(\mathbf{x}_{1:T}|\mathbf{x}_{0})} \right] = \mathbb{E}_{q} \left[ \log p_{\theta}(\mathbf{x}_{0:T}) - \ \log q(\mathbf{x}_{1:T}|\mathbf{x}_{0}) \right]
\end{equation} 

\section{Method}
\subsection{Disentangled Diffusion Autoencoders} 
Producing synthetic brain images with fine-grained control over conditional information such as site can be difficult due to the entanglement of these features with unknown biological variance. To resolve this, we developed a new class of diffusion model, called the disentangled diffusion autoencoder (DDAE). The DDAE, extends the diffusion autoencoder \cite{DiffAE} framework by allowing for known conditional variables, $\boldsymbol{c}$, to have a latent $f_{\psi}(\boldsymbol{c}) = \mathbf{z}_{\kappa}$. Here, the subscript $\kappa$ denotes that this variance is known. We also have a separate latent $s_{\phi}(\mathbf{x}_{0}) = \mathbf{z}_{\upsilon}$, where the subscript $\upsilon$ denotes that this variance is unknown.  Where ($\mathbf{z}_{\kappa}, \mathbf{z}_{\upsilon}) \in \mathbb{R}^{d}$. Under this formulation the forward process $q(\mathbf{x}_{1:T}|\mathbf{x}_{0})$ models the stochastic information within the image. Whilst $\mathbf{z}_{\kappa}$ models the known semantics and $\mathbf{z}_{\upsilon}$ models the unknown semantics. This approach produces representations that are semantically rich and disentangled, allowing for better control when generating conditional synthetic images. Our approach admits the following modified reverse process: 
\begin{equation}
    p(\mathbf{x}_{0:T}| \mathbf{z_{\kappa}, \mathbf{z}_{\upsilon}}) = p(\mathbf{x}_{T}) \prod_{t=1}^{t=T}p_{\theta}^{(t)}(\mathbf{x}_{t-1}|\mathbf{x}_{t}, \mathbf{z_{\kappa}, \mathbf{z}_{\upsilon}})
\end{equation} 


\subsection{Training Objective}
Our method makes use of the reparameterisation trick, after $t$ noising steps our noisy image $\mathbf{x}_{t}$ can be expressed as $\mathbf{x}_{t} = \sqrt{\alpha_{t}}\mathbf{x}_{0} + \sqrt{1 - \alpha_{t}}\epsilon$ where $\epsilon \sim \mathcal{N}(\mathbf{0}, \mathbf{I})$ and $\alpha_{t} = \prod_{s=1}^{t}(1 - \beta_{s})$ \cite{DDIM}. As such, instead of maximising a lower bound on log likelihood as in equation \ref{lower-bound}, we sample a batch of timesteps, $t$, apply $\mathbf{x}_{t} = \sqrt{\alpha_{t}}\mathbf{x}_{0} + \sqrt{1 - \alpha_{t}}\epsilon$ to our data, $\mathbf{x}_{0}$ and use a network $\epsilon_{\theta}(\mathbf{x}_{t}, \mathbf{z}_{\kappa}, \mathbf{z}_{\upsilon})$ to predict the $\epsilon$ that was sampled when noising the image. This admits the following objective:
\begin{equation}
    \label{Equation-4}
    \mathcal{L} := \sum_{t=1}^{T}\mathbb{E}_{\mathbf{x}_{0} \sim q(\mathbf{x}_{0}), \epsilon_{t} \sim \mathcal{N}(\mathbf{0}, \mathbf{I})} \left[ \| \epsilon_{\theta}^{(t)}(\mathbf{x}_{t}, \mathbf{z}_{\kappa}, \mathbf{z}_{\upsilon}) - \epsilon_{t} \|^{2}_{2}  \right] 
\end{equation}

See figure \ref{ModelArch} for our full model pipeline. We compare our proposed model to three baselines; ComBat, a cVAE, and Style-Encoding GAN \textemdash a statistical method, and two deep generative models, respectively. 
\begin{figure}[t]
\centering
    \includegraphics[width=\textwidth]{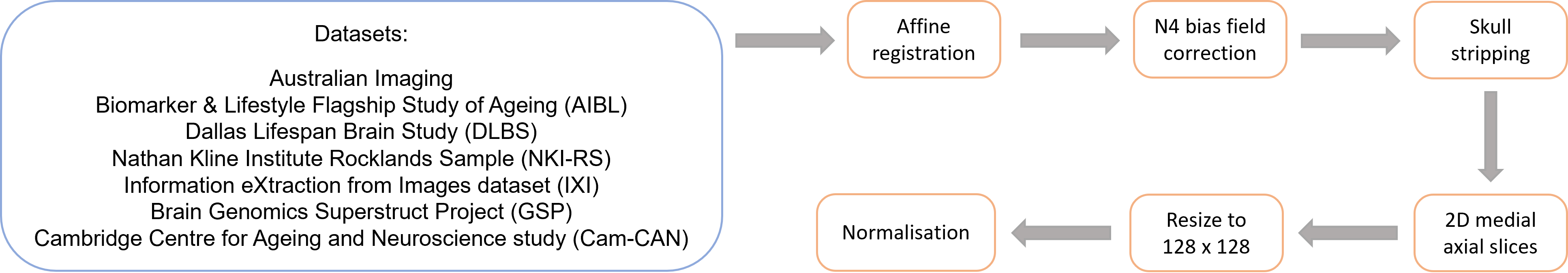}
\caption{Data pre-processing steps. The ANTS package was used to conduct affine registrations of the images to the MNI 152 brain template, they were then resampled to 130 × 130 × 130 resolution, Simple ITK was used to perform n4 bias field correction and HD-BET was used to skull strip the images. Following pre-processing, 2D medial axial slices were taken from the 3D volumes, which had their pixel values normalised to be between 0 and 1, after being resized to 128 × 128 resolution. The 2D images were then used to train the harmonization models.} 
\label{preprop}
\end{figure}  
\section{Experiments}
\subsection{Datasets}
Our dataset consisted of 4120 2D structural T1-weighted MR images from healthy controls (mean age = 41.7 years, std = 21.7 years). The data was drawn from 7 publicly available datasets, these were: the Australian Imaging Biomarker \& Lifestyle Flagship Study of Ageing (AIBL); the Dallas Lifespan Brain Study (DLBS); the Nathan Kline Institue Rocklands Sample (NKI-RS); Information eXtraction from Images dataset (IXI); the Brain Genomics Superstruct Project (GSP) and the Cambrdige Centre for Ageing and Neuroscience study (Cam-CAN). 
\subsection{Pre-processing}
Figure \ref{preprop} displays our pre-processing pipeline. The ANTS package was used to conduct affine registrations of the images to the MNI 152 brain template, they were then resampled to 130 × 130 × 130 resolution, Simple ITK was used to perform n4 bias field correction and HD-BET was used to skull strip the images. Following pre-processing, 2D medial axial slices were taken from the 3D volumes, which had their pixel values normalised to be between 0 and 1, after being resized to 128 × 128 resolution. The 2D images were then used to train the harmonization models. 
\begin{figure*}[t]
\centering
    \includegraphics[scale=0.20]{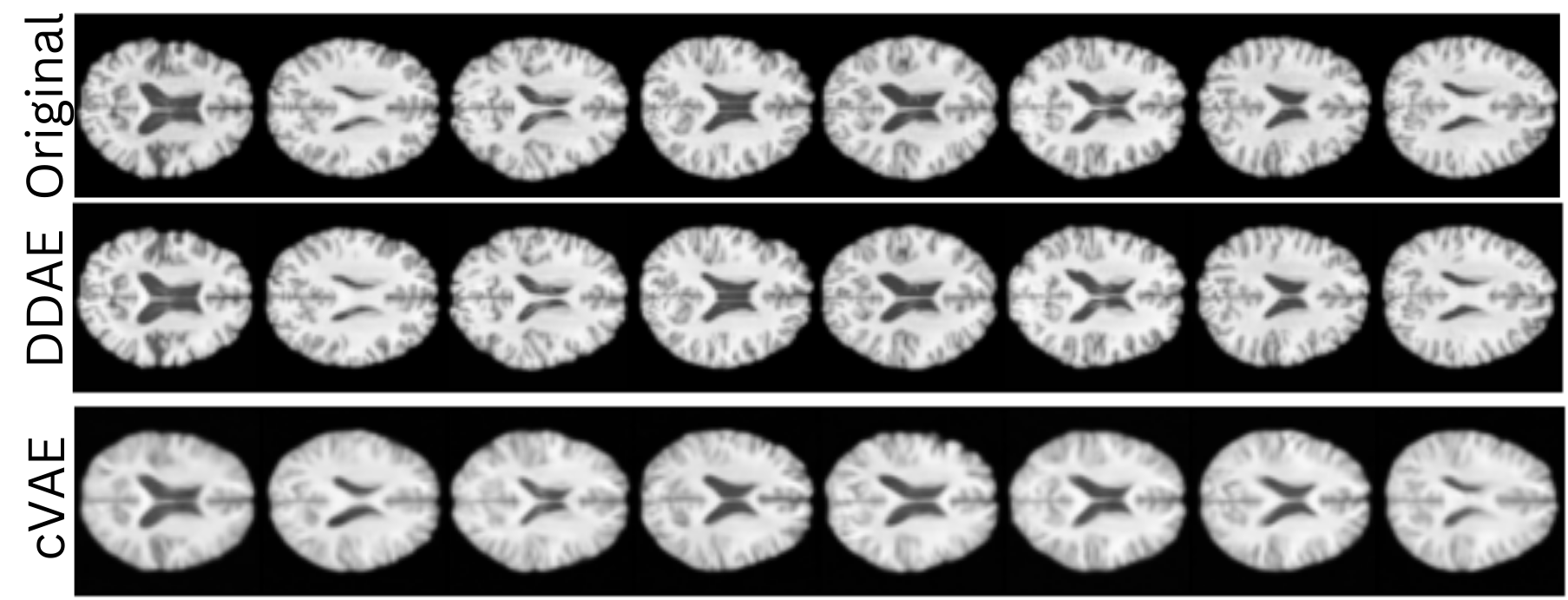}
\caption{Example Reconstructions from our Disentengled diffusion autoencoder (DDAE) compared to a cVAE. The first row shows original images from our dataset, the second row shows reconstructions from our model and the third row shows reconstructions from a cVAE.} 
\label{SmallRecons}
\end{figure*}  

\begin{figure}[t]
    \centering 
    \includegraphics[width=0.95\textwidth]{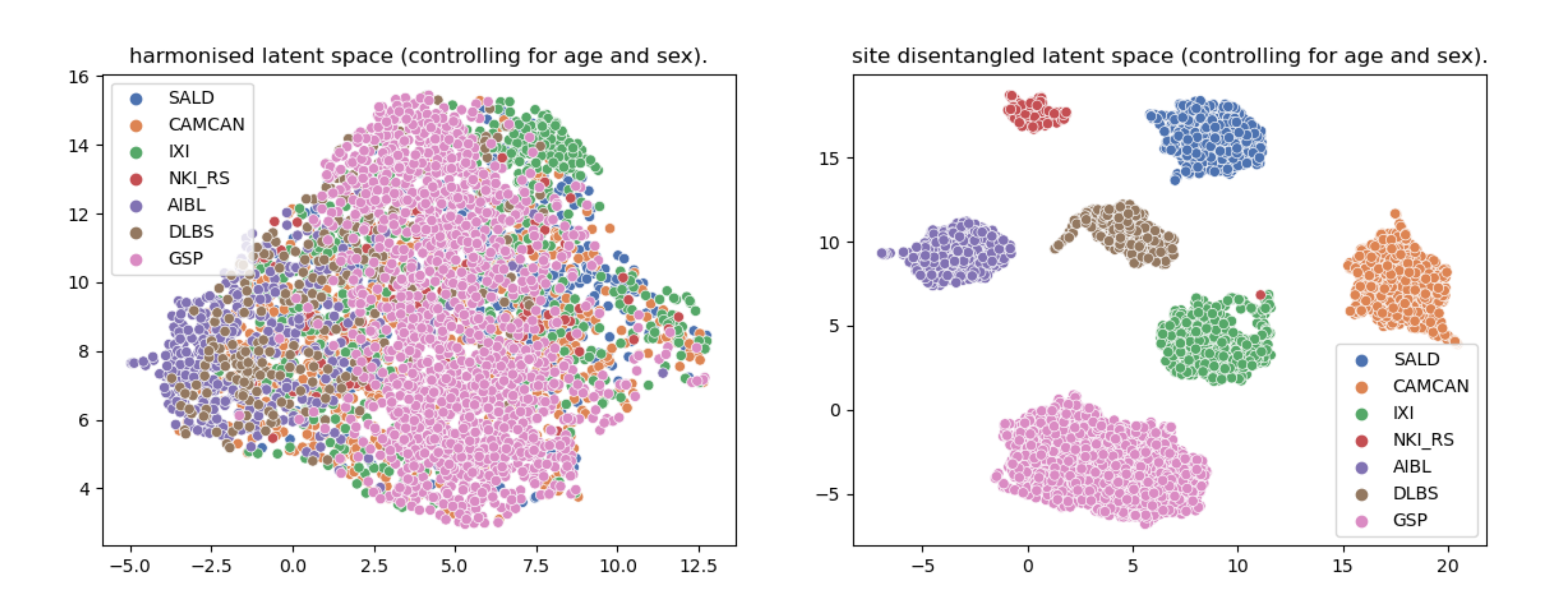}
    \caption{UMAP embeddings of the joint latent space of $\mathbf{z}_{\kappa}$ and $\mathbf{z}_{\upsilon}$. The image on the left shows the latent space when we harmonise the images by setting their site to the same dataset (IXI). The right image shows the disentangled latent space when we use the actual site labels. In both instances, we control for sex and age.}
    \label{fig:umaps}
\end{figure}
\subsection{Training setup and Benchmarking}
We trained the DDAE, ComBat, cVAE, and Style-Encoding GAN models using the full dataset of 4120 images. The noise predictor of the DDAE model, $\epsilon_{\theta}(\mathbf{x}_{0}, \mathbf{z}_{\kappa}, \mathbf{z}_{\upsilon})$ shared the U-Net architecture of Preechakul et al. (2022) \cite{DiffAE}. To train the ComBat model, we used the publicly available code from the NeuroComBat paper \cite{Fortin2018} with age and sex as biological covariates. The cVAE model was parameter matched to our proposed DDAE model and shared the same architecture as our noise predictor, omitting skip connections. Implementation details of the DDAE and Style-Encoding GAN are given in the Appendix. For DDAE, cVAE, and ComBat to generate the harmonized dataset, we mapped the unharmonized images to the IXI site, as this site had the greatest number of samples. For the Style-Encoding GAN, we mapped all images to a default reference image. To assess model performance, we used CNN classifiers and regressors to predict site label and biological covariates from the unharmonized and harmonized images. Full details of the model settings and network architectures are provided in the Appendix. All training was conducted on an NVIDIA RTX 4090. 
\begin{figure*}[t]
\centering
    \includegraphics[scale=0.7]{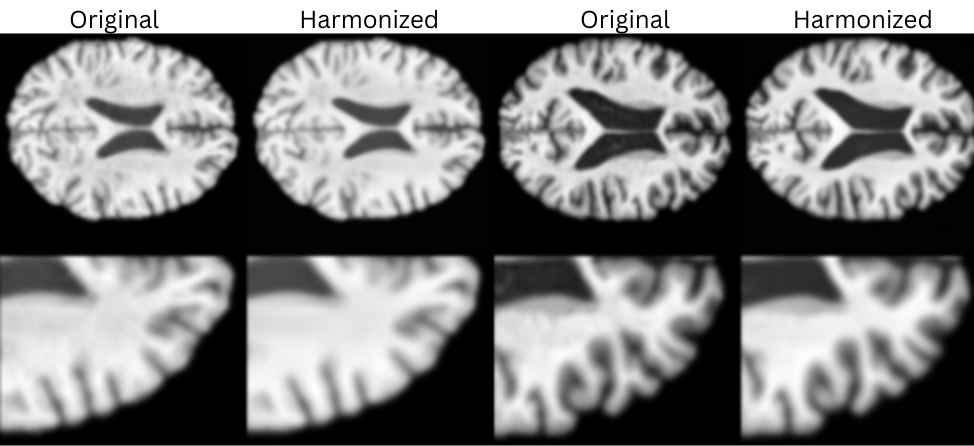}
\caption{Example images following harmonization. The first and third columns display images from two different datasets, and the second and fourth rows display the images after being harmonized. It can be seen that the images have their pixel intensity values increased accordingly whilst maintaining their original neuroanatomy. } 
\label{fig:harmony}
\end{figure*} 
\subsection{Results}
We assessed the performance of our model across 4 sub-tasks: ability to generate high quality images, remove site effects, preserve biological variance, and preserve individual differences from sources of unknown variance. Table \ref{tab:quantiative_results} provides an overview of the quantitative performance of all tested models. 

\subsubsection{Generating high-quality images.} For cVAE and DDAE, we reconstructed the original image for each sample without modifying the site label. Figure \ref{SmallRecons} shows a selection of generated images for the cVAE and DDAE models, where we can see visually that the DDAE gives better results with the cVAE model producing blurry images. To quantitatively evaluate the quality of the generated images and the ability of the proposed method to preserve the integrity of the image, we calculated the fretchet-inception distance (FID) \cite{fid} score between the original and site adjusted generated images for each model (see Table \ref{tab:quantiative_results}). The DDAE model achieved the lowest FID score, further illustrating that our model can generate higher-quality images than previous approaches. 

\subsubsection{Removing site effects.} To validate that our model can disentangle and remove site effects, we first generated UMAP plots from the joint latent space of $\mathbf{z}_{\kappa}$ and $\mathbf{z}_{\upsilon}$ under two conditions. First, we set the site effect to the true label (disentanglement of site effects), then we set all the sites to the same label (removal of site effects). As seen in Figure \ref{fig:umaps}, when conditioning on the true site labels the latent representations are clearly disentangled. In contrast, when the sites are set to be equal, the DDAE latent vectors exhibit greater homogeneity, indicating successful disentanglement and removal of site effects. Next, we assessed the ability of DDAE to remove site effects in the data space. Figure \ref{fig:harmony} shows example reference images, harmonized to the IXI site using DDAE. The harmonized images appear more homogeneous with respect to image intensity while individual factors such as ventricular volume and sulcal morphology are preserved. To quantify the ability of each harmonization method to remove site effects, we trained a CNN classifier (as described in the Appendix) to predict site. The classification accuracies, averaged across 5 seeds, on the test set are given in Table \ref{tab:quantiative_results}. The DDAE significantly outperforms ComBat which preserves site-identifiable information. Interestingly, the CNN classifier achieves a considerably higher accuracy in predicting site when using the ComBat harmonized data compared to using the unharmonized data. Zhao et al. (2023) \cite{Zhao2023} previously observed high site label prediction accuracy using ComBat harmonized vertex-wise imaging data, which highlights the need for improved harmonized methods. Although the cVAE and Style-Encoder GAN generated images effectively remove site effects, this comes at the cost of image quality and biologically salient information, which we discuss below.

\begin{table}[t]
\centering 
\caption{Overview of quantitative test results. We evaluated the ability of models to preserve known (age, sex) variance, and unknown (PCC) variance, as well as produce high-quality images (FID), whilst removing site effects.}\label{tab:quantiative_results}
\begin{tabular}{|l|l|l|l|l|l|}
\hline
Metric & FID $\downarrow$ & Site Acc $\downarrow$ & Age $R^2$ $\uparrow$ & Sex Acc $\uparrow$ & PCC $\uparrow$ \\
\hline
\textbf{DDAE} & $7.40$ & $58.56\pm3.00$ & $0.810\pm0.0141$ & $72.18\pm2.16$ & $0.982$ \\
\hline
ComBat & $13.43$ & $89.07\pm18.58$ & $0.856\pm0.0104$ & $73.23\pm1.67$ & $0.972$ \\
\hline
cVAE & $73.60$ & $23.04\pm1.14$ & $0.441\pm0.0355$ & $95.41\pm2.11$ & $0.228$ \\
\hline
Style-Encoder GAN & $42.65$ & $49.90\pm4.94$ & $0.773\pm0.713$ & $65.58\pm3.42$ & $0.884$ \\
\hline
unharmonized & $-$ & $61.17\pm8.95$ & $0.832\pm0.0282$ & $70.24\pm1.22$ & $-$ \\
\hline
\end{tabular}
\end{table}
\subsubsection{Predicting known biological covariates.} We assessed the ability to predict age from the harmonized or unharmonized images using a CNN regressor. The coefficient of determination ($R^2$) values on the test set are given in Table \ref{tab:quantiative_results}. DDAE displays comparable performance to both ComBat and the unharmonized data, which highlights the ability of our proposed model to preserve age-related information. Conversely, the cVAE and GAN struggle to retain age information. Similarly, we trained a CNN classifier to predict sex. The test set results are given in Table \ref{tab:quantiative_results}. Again, our proposed model produces results that are comparable to ComBat, and the classifier was better able to predict sex from the DDAE harmonized images than the unharmonized images. We note that the cVAE model generates images from which sex can be predicted with a high degree of accuracy, but this underscores a limitation in the model's ability to finely control the generation of features. Among the metrics we employed, information related to sex is the only feature retained by the cVAE, as detailed in Table \ref{tab:quantiative_results}. Thus, despite the site classifier results indicating better site removal for the cVAE and Style-Encoder GAN models, they weren't in general as effective at preserving biologically relevant information compared to DDAE. As such, these results highlight the ability of our model to remove site effects whilst preserving biological variance.

\subsubsection{Preserving within-site variability.} We followed the procedure introduced by Zhao et al. (2019) \cite{Zhao2019}. For each site, we calculated the Euclidean distances between all pairs of images, forming a distance matrix; $D_{i, j}^{n \times n}=\left\|x_i-x_j\right\|^2$ where $n$ is the number of images for a specific site. We then calculated the Pearson correlation coefficient (PCC) between $D_{i, j}^{n \times n}$ for the unharmonized data and $D_{i, j}^{n \times n}$ for the harmonized data from each harmonization method (Table \ref{tab:quantiative_results}).  Our proposed model gave the highest PCC, suggesting that it is best able to preserve individual differences. 

\begin{table}
\centering 
\caption{An overview of each model's performance across the four main goals of data harmonization, as measured by our quantitative tests.}\label{tab:overall_performance}
\begin{tabular}{|l|l|l|l|l|}
\hline
Metric & Known Variance  & Unknown Variance    & Image quality   & Site Effects \\
\hline 
\textbf{DDAE} & \checkmark  & \checkmark   & \checkmark   & \checkmark \\
\hline
ComBat  & \checkmark  & \checkmark & $\times$  & $\times$ \\
\hline
cVAE  & (\checkmark)  &  $\times$   & $\times$  & \checkmark \\
\hline 
Style-Encoder GAN   & (\checkmark)  & $\times$   & $\times$ & \checkmark \\
\hline 
\end{tabular}
\end{table} 

Table \ref{tab:overall_performance} summarizes each model's performance across the four sub-tasks. While the cVAE and Style-Encoder GAN successfully remove site effects and retain some biological variance, they struggle to preserve unknown variance and produce high-quality images. ComBat, although generally capable of preserving biological variance in our dataset, fails to effectively eliminate site effects. The DDAE stands out as the only model performing well across all four tasks.

\section{Conclusions and further work}
In this work, we proposed a novel class of diffusion model, the DDAE, which we apply to the task of harmonizing high-dimensional neuroimaging data. Our proposed model builds upon recent advances in diffusion models to generate high-quality images with specific characteristics. Compared to ComBat, a popular site removal technique, our proposed model was better able to remove site effects while preserving biological variability. Whilst recently proposed generative models for site removal are able to remove site-identifiable information, they do so at the expense of removing biologically relevant information and generate lower-quality images where individual-level differences have been erased. As such, we have shown both through visual and quantitative results that our model is overall the best equipped for removing site effects, preserving unknown and known variability, and generating high-quality images compared to previous approaches. Furthermore, the diffusion model presented here is not limited to multi-site data harmonization, and the methodological developments could be adapted for other applications, both within and outside of the medical domain. A limitation of our proposed model is that it is designed for 2D images. Further work will focus on extending our proposed model to vertex-wise data. 
\bibliographystyle{}

\section{Appendix}
\begin{table}
\centering 
\caption{DDAE model elements and architectures.  For the age, site and sex predictor networks, a ReLU, batch normalisation and $0.5$ dropout was applied between layers. There were $123$ million parameters in total. The model was trained on an Nvidia rtx 4090 graphics card.}
\begin{tabular}{|l|l|}
\hline
Model element &  Architecture\\
\hline 
Noise predictor network, $\mathbf{\epsilon}^{(t)}_{\theta}(\mathbf{x}_{0}, \mathbf{z}_{\text{sem}})$ & U-net \\
\hline
Unknown variance encoder, $\mathbf{s}_{\phi}(\mathbf{x}_{0})$   & Downward path and middle block of U-net\\
\hline 
Known variance encoder $f_{\psi}(\textbf{c})$  & MLP layers (3 → 1526 → 512 )\\ 
\hline
Age predictor network  & MLP layers (512 → 128 → 32)   \\
\hline 
Site predictor network  &  MLP layers (512 → 128 → 32)   \\
\hline 
Sex predictor network  &  MLP layers (512 → 128 → 32)   \\
\hline 
\end{tabular}
\end{table}

\begin{figure*}
\centering
    \includegraphics[width=\textwidth]{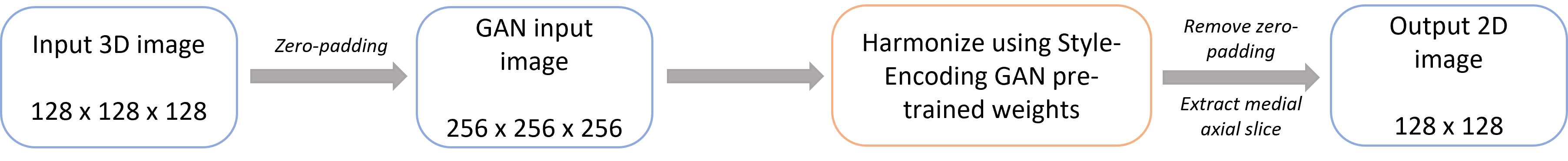}
\caption{Framework for harmonizing images using Style-Encoding GAN using their provided code \url{https://github.com/USC-LoBeS/style_transfer_harmonization}. Pre-processed images were zero-padded prior to using the provided pre-trained model weights with the default parameters. From the model outputs we selected the final harmonized and intensity re-scaled images that had undergone artifact removal. We ran this model on an Nvidia A100 GPU. To carry out downstream analyses, we removed the zero-padding on the harmonized images, extracted the 2D medial axial slices, and resized them to a resolution in keeping with the outputs of our model.} 
\end{figure*}  

\begin{figure*}
\centering
    \includegraphics[width=0.9\textwidth]{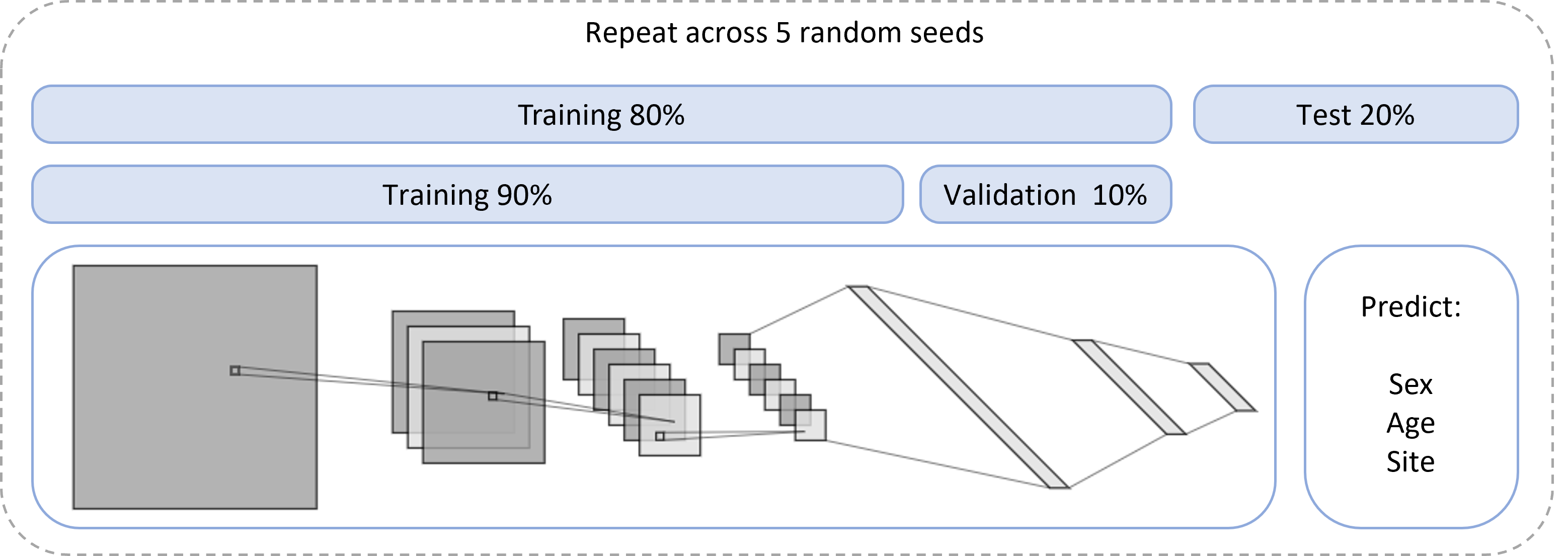}
\caption{Framework for the CNN classifiers and regressors. Images underwent 3 convolutional layers (kernel size: 4, stride: 2, padding: 1), followed by a MLP with three layers (1536 → 768 → 128), applying ReLU functions between layers. For sex and age prediction, the final MLP layer yielded a single output, while for site, it generated a vector of size 7. The training set was split into training/validation sets, employing early stopping based on validation loss. Network weights were optimized using an Adam optimizer (learning rate: 0.0001) with objectives: BCE loss for sex, MSE loss for age, and CE loss for site.} 
\end{figure*}  
\end{document}